%% file: iclr2023_conference.tex
\documentclass{article} 
\usepackage{iclr2023_conference,times}

\input{math_commands.tex}

\usepackage{hyperref}
\hypersetup{
    colorlinks=true,
    linkcolor=magenta,
    filecolor=magenta,      
    urlcolor=magenta
    }
\usepackage{url}
\usepackage{caption, subcaption, floatrow}

\title{About latent roles in forecasting players in team sports}

\author{Luca Scofano, Alessio Sampieri, Giuseppe Re, Matteo Almanza \\ \textbf{Alessandro Panconesi, Fabio Galasso} \\
Sapienza University of Rome\\
\texttt{\{scofano,sampieri\}@diag.uniroma1.it} \\
\texttt{\{re,almanza,ale,galasso\}@di.uniroma1.it} \\
}

\iclrfinalcopy 
\begin{document}

\maketitle

\input{src/sections/0.titleAbstract.tex}

\input{src/sections/1.introduction.tex}

\input{src/sections/2.related_work.tex}

\input{src/sections/3.methodology.tex}

\input{src/sections/4.experimental_evaluation.tex}

\input{src/sections/5.conclusion.tex}

\bibliography{iclr2023_conference}
\bibliographystyle{iclr2023_conference}


\end{document}

%% file: math_commands.tex
\usepackage{amsmath,amsfonts,bm}

\def\eqref#1{equation~\ref{#1}}

\def\1{\bm{1}}

\DeclareMathAlphabet{\mathsfit}{\encodingdefault}{\sfdefault}{m}{sl}
\SetMathAlphabet{\mathsfit}{bold}{\encodingdefault}{\sfdefault}{bx}{n}



%% file: src/sections/0.titleAbstract.tex
\begin{abstract}
Forecasting players in sports has grown in popularity due to the potential for a tactical advantage and the applicability of such research to multi-agent interaction systems. Team sports contain a significant social component that influences interactions between teammates and opponents. However, it still needs to be fully exploited.
In this work, we hypothesize that each participant has a specific function in each action and that role-based interaction is critical for predicting players' future moves. We create \emph{RolFor}, a novel end-to-end model for Role-based Forecasting. RolFor uses a new module we developed called Ordering Neural Networks (OrderNN) to permute the order of the players such that each player is assigned to a latent role.
The latent role is then modeled with a RoleGCN. Thanks to its graph representation, it provides a fully learnable adjacency matrix that captures the relationships between roles and is subsequently used to forecast the players' future trajectories.
Extensive experiments on a challenging NBA basketball dataset back up the importance of roles and justify our goal of modeling them using optimizable models.
When an oracle provides roles, the proposed RolFor compares favorably to the current state-of-the-art (it ranks first in terms of ADE and second in terms of FDE errors).
However, training the end-to-end RolFor incurs the issues of differentiability of permutation methods, which we experimentally review. Finally, this work restates differentiable ranking as a difficult open problem and its great potential in conjunction with graph-based interaction models. Project is available at: \href{https://www.pinlab.org/aboutlatentroles}{https://www.pinlab.org/aboutlatentroles}
\end{abstract}

%% file: src/sections/1.introduction.tex
\section{Introduction}
\begin{figure*}[!htp]
    \centering
        \includegraphics[scale =0.50]{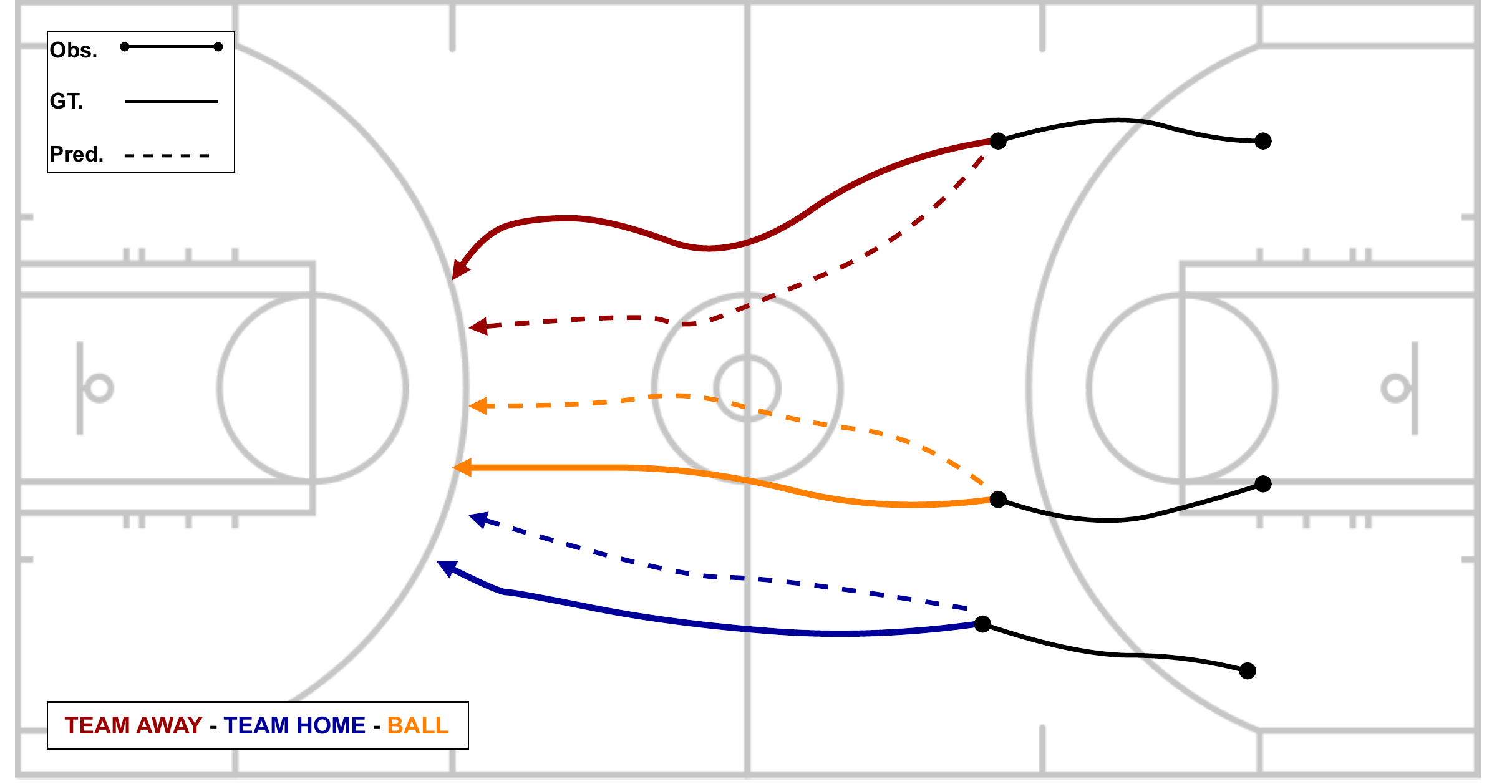}
    
    \caption{Example of multi-agent trajectory forecasting. We only plot one player for each team and the basketball for readability reasons.}
    \label{teaser}
\end{figure*}

Recent advances in visual recognition and sequence modeling have enabled novel objectives in athletic performance and sport analytics~\cite{rein2016big,merhej2021happened,morgulev2018sports}. One novel and challenging task is the multi-agent trajectory forecasting (See Fig. \ref{teaser}) of the players as a result of their observed current motion~\cite{evolvegraph, multimodalnba}. The difficulty is due to tactics, tight interaction of team players, the antagonist behavior of opponents, and the role assigned to each player in each action.
Traditional trajectory forecasting techniques~\cite{sociallstm, transformer, stgat, socialgan, socialstgcn} fall short in performance due to their general formulations and lack of sport-specific dynamics. Furthermore, trajectory forecasting methods must deal with the variable numbers of people in each scene (usually absent in games) and do not consider the presence of two opposing teams, the ball, or the finality in the given sport (e.g.~scoring). Most recent literature~~\cite{evolvegraph, multimodalnba} has started to address some of these objectives, but, to our knowledge, none has modeled the role of players for specific actions. 

We propose RolFor, a novel graph-based encoder-decoder model that performs a robust prediction of the players' future trajectory, utilizing roles to comprehend their interactions. The players' positions and movements on the court often follow pre-defined schemes, so we assume that each player may be assigned a specific role. By proposing a role-based ordering of nodes in the graph, it is possible to establish a player order and learn role-specific relationships.

The current best performers in-game forecasting~\cite{evolvegraph, socialstgcn} are based on graph convolutional networks (GCN)~\cite{kipfgcn}, but they do not consider roles. On the contrary,  we model latent roles as nodes in the graph.
Our RolFor model is composed of an ordering and a relational module. The former is an Ordering Network, which identifies latent roles and orders players according to them -- we use a well-known sorting approximation \cite{softrank} to order the latent projections of the players. In the latter, the game dynamics and trajectories are modeled using RoleGCN, based on \cite{stsgcn} where the nodes are the newly assigned roles, and the edges are their relations. The adjacency matrix is learned, and each entry corresponds to learning the role-based player interaction. 

We assume roles exist, and many characteristics could dictate them -- e.g., marking the opponent, possessing the ball, and identifying the attacking and defending teams. However, we assume no prior knowledge
about roles.
Our goal is to learn latent roles with an end-to-end algorithm, only considering the future trajectory of all players.
To test our intuition about roles, we pre-processed the basketball dataset by assigning roles based on different methods (Table \ref{tab:oracle_orderings}) and using those in our RoleGCN. We produce SOTA results, confirming that finding good roles improves model performance.
Nevertheless, we found that current differentiable ordering methods face some limitations of backpropagation when inserted in complex models. 
In summary, our contributions are:
\begin{itemize}
    \item We experimentally demonstrate that leveraging roles yields SoTA in trajectory forecasting.
    \item We propose an Order Neural Network module that creates a latent representation of the player's coordinates and orders them accordingly.
    \item We build a RoleGCN that learns the relations among roles.
    \item We empirically demonstrate that the current differentiable ordering approaches have some difficulties with backpropagation -- enabling little to no gradients to flow through -- when dealing with complex models.
\end{itemize}

%% file: src/sections/2.related_work.tex
\section{Related Work}

\paragraph{Trajectory Forecasting} 
The forecasting of pedestrian movement has been studied to deal with realistic crowd simulation~\cite{pelechano07} or to improve vehicle collision avoidance~\cite{bhattacharyya2018long}; it was also used to enhance the accuracy of tracking systems \cite{choi12, pellegrini10, yamaguchi11} and to study the intentions of individuals or groups of people \cite{lan12,xie18}.
Different models have been proposed to predict such trajectories, like Long Short-Term Memory (LSTM) networks \cite{lstm} with shared hidden states \cite{sociallstm}, multi-modal Generative Adversarial Networks (GANs) \cite{socialgan}, or inverse reinforcement learning   \cite{activityfore}.
This group forecasting scenario resembles Game Forecasting, where it is necessary to model the movements of two opposing teams.

\paragraph{Game forecasting}
Associations such as National Basketball League or the English Premier League have used sophisticated tracking systems that allow teams to gain insight into each game \cite{carling2008role}. Variational Autoencoders (VAEs) were used to model real-world basketball actions, showing that the offensive player trajectories are less predictable than the defense\cite{felsen2018will}. LSTM\cite{seidl10} were employed to predict near-optimal defensive positions for soccer and basketball, respectively, as for predicting the player's movements during the game \cite{multimodalnba}. Variants of VAEs have also been used \cite{sun18} to generate trajectories for NBA players.
NBA player trajectory forecasting was also studied in \cite{zhan18} and \cite{zheng17}, proposing a deep generative model based on VAE, LSTM, and RNN \cite{multimodalnba, lstm, rnn} and trained with weak supervision to predict trajectories for an entire team. 
Nonetheless, we did not encounter work estimating specific latent roles and learning the player interaction on those bases.

\paragraph{GCN-based forecasting}
Adopting a graph structure makes it possible to encode information and quantify shared information between nodes. SoA in pose forecasting learns specific terms for the specific joint-to-joint relation \cite{stsgcn, stgcn}.
Graphs are also widely used in trajectory forecasting and can be considered fully connected \cite{evolvegraph}, sparse or weighted. These structures distinctly model the interrelationships between nodes, and their combination can be crucial. Also, Graph attention layers (GAT) are widely used in trajectory forecasting \cite{stgat, li2021grin} to learn the inter-player dependencies. 
We use the SoA pose forecasting model \cite{stsgcn} to model role-based interaction. Pose forecasting is relevant since it considers the fixed node cardinalities and the learned interactions. However, players from various matches and teams do not have a fixed order, which is not an issue with pose forecasting. This encourages us to learn and re-order the players based on hidden roles.

\paragraph{Differentiable Ranking} 
Sorting and Ranking are two popular operations in information retrieval that, in our case, can be useful in identifying the role of players. 
In composition with other functions, sorting induces non-convexity, rendering model parameter optimization difficult.
On the other hand, the ranking operation outputs the positions, or ranks, of the input values in the sorted vector. Gradient computation is far more complicated as a piece-wise constant function and could prevent gradient backpropagation.
Several recent works \cite{cuturi19, softrank} provide an approximation of the above operations for use in a learnable framework.

%% file: src/sections/3.methodology.tex
\section{Methodology}

This section formally defines the problem and explains our strategy to tackle it, focusing on the role assignment and encoding methods.
First, we briefly explain how the Role-based Forecasting model (RolFor) performs latent mapping, role assignment, and trajectory prediction.
We also focus on the main components: the Order Neural Network (OrderNN), which handles the ordering task, and the RoleGCN, which facilitates the learning process of relationships between roles in a game.

\subsection{Problem Formalization}
We target to predict the future trajectory of all players, given the observed positions at past time frames. 
We denote the players by 2D vectors $x_{p, t}$ representing player $p$ at time $t$. The position of all players at time $t$ are aggregated into a matrix of 2D coordinates $X_t \in \mathbb{R}^{2 \times p}$. Motion history of players is denoted by the tensor $X_{in} = [X_1, X_2, ..., X_T]$, which is constructed out of the matrices $X_t$  for frames $t = 1, ...,  T$. 
The goal is to predict the future $K$ players positions $X_{out} = [X_{T+1}, ..., X_{T+K}]$. 

\subsection{Role-based Forecasting model (RolFor)}
RolFor uses two main components, the first one being the OrderNN (Section 3.2.1), which orders players according to their latent roles. We postulate the existence of latent roles that when learned in an end-to-end architecture yield the best trajectory forecasting performance. From the OrderNN, we will consider $R$, the role vector, instead of $P$, the position vector. Notice that $R$ and $P$ have equal dimensions.
The graph is now defined as $\mathcal{G}=(\mathcal{V}, \mathcal{E})$, where the nodes indicate the roles of each player and the edges capture the interaction among roles during the game. The graph $\mathcal{G}$ has $|\mathcal{V}| = T \times R$ nodes, which represent all $R$ roles across $T$ observed time frames. Edges in $\mathcal{E}$ are represented by a Spatio-Temporal adjacency matrix $A^{st} \in \mathbb{R}^{RT \times RT}$, relating the interactions of all roles at all times. Note that $A^{st}$ is learned, i.e., \ the model learns how players with different roles interact by learning how latent roles interact over time. 

\subsubsection{Order Neural Network}
\begin{center}
\begin{figure}[!htp]
    \centering
    \includegraphics[scale=0.9]{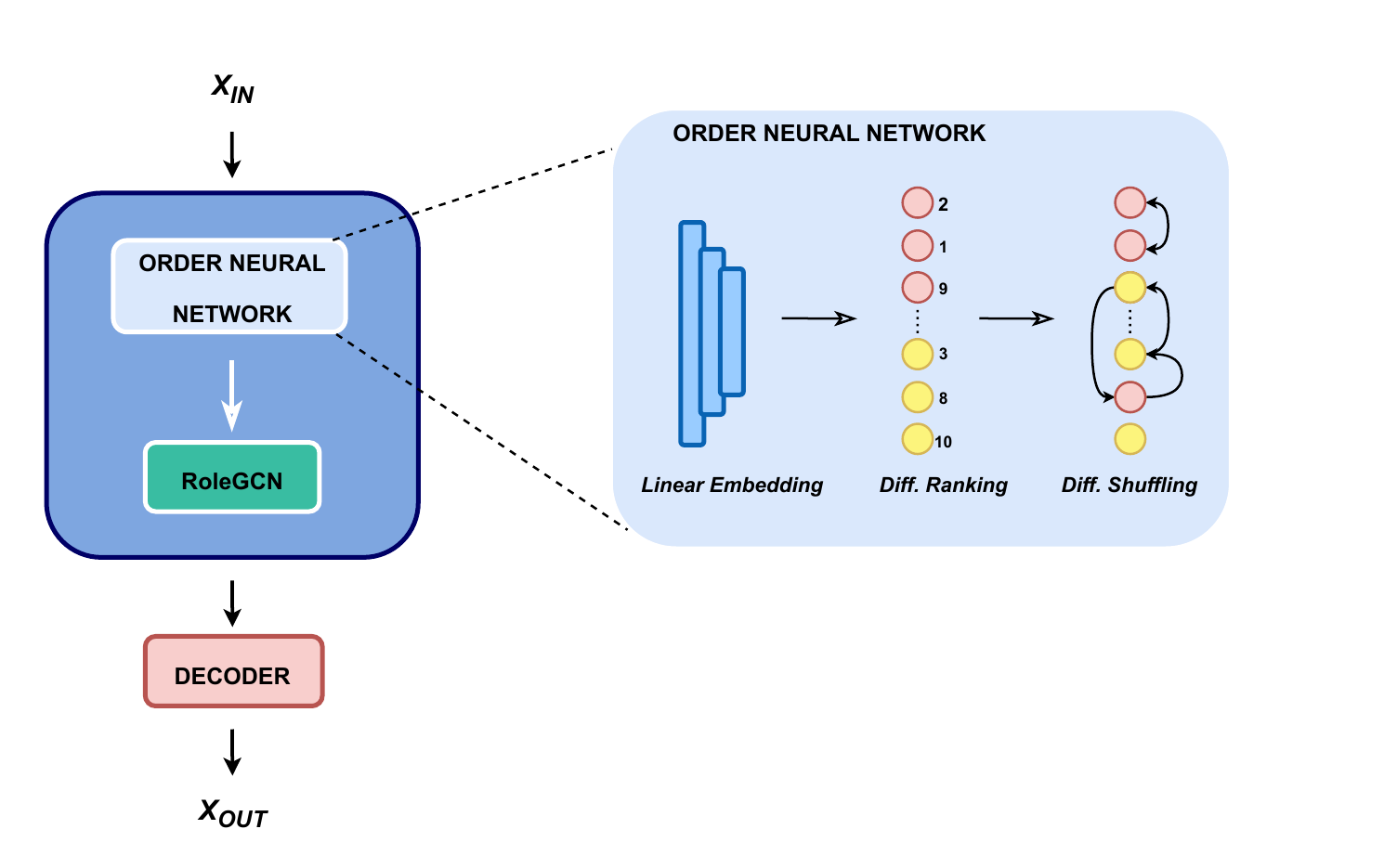}
     
    \caption{Architecture of RolFor and a zoom into Order Neural Network}\label{fig:network}
    
\end{figure}
\end{center}

The Order Neural Network (Fig.~\ref{fig:network}) takes in input the initial coordinates $X_{in}$ and maps them into a latent space. Additionally,  it orders the latent vector into optimal roles $X_{role\_in}$, thanks to the use of a differentiable ranking method \cite{softrank}, which has the same dimensionality of $X_{in}$. Note that roles get the corresponding position coordinates over subsequent time frames, so each role is now characterized by a spatio-temporal trajectory.
A straightforward example of a role assignment involves sorting players in ascending order based on their Euclidean distance from the ball.
This method is also used as a valuable proxy task, which we use for ablation studies (see Section \ref{4.experiments} Table \ref{tab:oracle_orderings}).
However, since RolFor is trained end-to-end, OrderNN is free to learn the ideal ordering that yields the best forecasting performance.

\paragraph{The differentiable ranking method} SoftRank, \cite{softrank} is a recent differentiable implementation of the classic sorting and ranking algorithm, empirically shown to achieve accurate approximation for both tasks. 
It is designed by constructing differentiable operators as projections onto the permutahedron, i.e., the convex hull of permutations, and using a reduction to isotonic optimization. 
The key takeaway of the method is to cast sorting and ranking operations as linear programs over the permutahedron. More precisely, it formulates the argsort and ranking operations as optimization problems over the set of permutations $\Sigma$. SoftRank also relies on a regularization parameter $\varepsilon$, which creates a trade-off between the differentiability of the algorithm and the optimum's accuracy. The greater the regularization factor ($\varepsilon  \rightarrow \infty$), the further the approximation from the permutation vertices, and the smoother the loss function gradient. And vice versa, by picking an $\varepsilon \rightarrow 0$, the algorithm will yield more accurate permutations with a lower degree of differentiability. 
After learning the ranking, we order the players according to it by employing a differentiable re-shuffling module.
The outputs of SoftRank are noted as $\left \{ s_i \right \}_{i=1}^{n}$ where \textit{n} is the number of rankings considered. At this point, we use a so-called \textit{base} matrix \textbf{B} with the number of rows and columns equal to the number of rankings. \textbf{B} will be used to store the real rankings $\left \{  p_i\right \}_{i=0}^n$. 
We then compute a $\left \{ \Delta_i \right \}_{i=1}^{n}$ matrix, which represents $\Delta_i = p_i-s_i$ for each position $\left \{  p_i\right \}_{i=0}^n$. The matrix $\Delta$ is used as the input as a rescaling function.
The re-shuffle process is a weighted combination: it yields a real shuffling when the approximated rankings are integer and a differentiable shuffling instead when the ranking is fractional. $M_i =  e^{(\frac{-\Delta}{scale})^2}$ can be considered an array of weights for each position, with values closer to 1 being the predicted positions of each player. Finally, this will be used to recall the initial coordinates in an ordered manner:




\begin{equation}
    P_i=\left\{\begin{matrix}x_i' = \sum_{j = 1}^{n} M_j \cdot x_j
    \\ y_i' = \sum_{j = 1}^{n} M_j \cdot y_j\end{matrix}\right.
\end{equation}

\subsubsection{RoleGCN}
Once the latent roles are inferred, the graph $\mathcal{G}=(\mathcal{V}, \mathcal{E})$ represents each node $i \in \mathcal{V}$ as the player's role while the edges $(i, r) \in \mathcal{E}$ connect all the roles and describe their mutual interaction. 
RoleGCN (Fig~\ref{fig:network}) will capture the underlying graph's relationships between different nodes on the court in the same time frame and between one node and itself over different time-frames.
GCN \cite{kipfgcn} is a graph-based operation that works with nodes and edges. For nodes, it aims to learn an embedding containing information about the node itself and its neighborhood for each node in the graph. 
Thus, the learned adjacency matrices yield a quantitative description of the interplay among roles. 
The space-time cross-talk is realized by factoring the space-time adjacency matrix (as in \cite{stsgcn}) into the product of separate spatial and temporal adjacency matrices $A^{st} = A^SA^t$.
A separable space-time graph convolutional layer \textit{l} is written as follows:

\begin{equation}\label{STSlayer}
    H^{(l+1)} = \sigma(A^{s-(l)} A^{t-(l)} \mathcal{H}^{(l)}W^{(l)})
\end{equation}

It is similar to a classic GCN convolutional layer, where $A^{s-(l)} A^{t-(l)}$ is the factorized matrix $A^{st-(l)}$ of a GCN \cite{kipfgcn} layer. The critical difference is better efficiency and allows full learnability of the former. 


\subsubsection{Decoder}
First, we de-shuffle the permuted roles according to the inverse of \textbf{B} to return to the original coordinates' position.
The decoding is done with multiple temporal convolutional (TCN) layers  \cite{convauto} used to predict the following frames.
We adopt TCN due to its performance and robustness.

%% file: src/sections/4.experimental_evaluation.tex
\section{Experimental Evaluation}\label{4.experiments}
In this section, we introduce the NBA benchmark dataset and metrics, the trajectory forecasting results and investigate why learning E2E roles is challenging.

\subsection{Dataset}
For our experiments, we use NBA SportVU \cite{felsen2018will}. It contains players and ball trajectories for 631 games from the 2015-2016 NBA season. Similar to previous work \cite{sun18}, we focus on just two teams and consider all their games.
We obtain a dataset of $95,002$, $12$-second sequences of players and ball overhead-view trajectories from $1247$ games. Each sequence is sampled at $25$ Hz, has the same team on offense for the entire duration and ends in a shot, turnover, or foul. As in \cite{felsen2018will}, the data is randomly split into train, validation, and test sets with respectively $60,708$, $15,244$, and $19,050$ sequences.

\subsection{Trajectory Forecasting Metrics}
We use as metrics ADE (Average Displacement Error) and FDE (Final Displacement Error), as usual in literature \cite{evolvegraph, multimodalnba, sociallstm, felsen2018will, socialgan}.
They are used to measure the error of the whole trajectory sequence and the final endpoints for each player.
Respectively:

\begin{equation}\label{ADE}
    ADE=\left \| \hat{T}_c-T_c \right \|_{2}^{2}
\end{equation}

\begin{equation}\label{FDE}
    FDE=\left \| \hat{E}_f-E_f \right \|_{2}^{2}
\end{equation}

Each observation has \textit{five} frames, corresponding to 2.0 seconds in a basketball scenario. The goal is to forecast the successive \textit{ten} frames (4.0 seconds). In Eq.~\ref{ADE}, $\hat{T}_c$ represents the prediction for all future trajectories over the $c = 1, .., 10$ subsequent frames, and $T_c$ is the ground truth. The same nomenclature is used in Eq.~\ref{FDE}, where $E$ is the matrix for the endpoints and $c = 1$ since we are only considering the last frame.

\subsection{Trajectory Forecasting Results}

\textit{So, do roles exist, and does learning the role interaction yields state-of-the-art performance?} We answer this question by considering the most straightforward ordering: Euclidean distance of players from the ball.
In Table~\ref{tab:generic_results}, we report state-of-the-art techniques compared to the RolFor model, with the Euclidean distance ordering of players from the ball. \cite{evolvegraph} proposes multiple predictions via latent interaction graphs among multiple interactive agents. \cite{socialgan}, similarly, is also a multi-modal model incorporating the social aspects of the players as well. \cite{stgat} is based on a sequence-to-sequence architecture to predict the future trajectories of players. Lastly, \cite{socialstgcn} substitutes the need for aggregation by modeling the interactions as a graph. Similar to \cite{stgcn}, it needs a pre-defined graph, allowing the leaning procedure only on the given edges.
RolFor in Table~\ref{tab:generic_results} yields the SoA forecasting performance in terms of ADE, 5.55 meters, second best in terms of FDE, 9.99 meters. It sorts players according to their Euclidean distance from the ball, arranging them into a sequence of attackers (players detaining the ball in the considered action), alternating with defenders (not detaining the ball). Each attacker is followed by its marker, which RolFor considers the closest to it in terms of Euclidean distance. As for all other reported SoA algorithms, RolFor considers that the teams are known. Finally, "Oracular Permutation'' means that RolFor uses distances at the last future step, i.e., \ step 10 in the future. In contrast, any other reported algorithm uses only the observed five frames. We will investigate this more thoroughly in the next section.
A neural network can learn the Euclidean distance, and softRank \cite{softrank} should be able to sort the players according to it. Replacing the hand-defined distance computation with a Neural Network should be as effective.
We expect that a model with a sorting unit that learns sorting E2E in relation to the final forecasting goal should be capable of doing better than this, assuming all modules are effectively differentiable.

\input{src/tables/comparison_existing.tex}

\subsubsection{Further Experiments on Euclidean Ordering} \label{sec:eucl-ord}
We delve deeper into the results of RolFor in Table~\ref{tab:generic_results} and analyze the importance of each hand-defined Euclidean distance term in Table~\ref{tab:oracle_orderings}.

\textit{No ordering Vs. Simple ordering.}
The first forecasting result in the table neglects the player ordering and learns interaction terms between players, arranged in random order. It yields 6.34/11.5 ADE/FDE meters errors.
Simple ordering stands for arranging all players in a list, according to their distance from the ball, at the last (5th) observed frame. This uncomplicated ordering is only negligibly better than no order. A GCN model may deal with players in random order well and only benefits from ordering if it is informative.

\textit{Distance from the ball and marking.}
Results in the third row of the Table~\ref{tab:oracle_orderings} add marking to the ball distance ordering. Each player in the attacker team is matched with one from the defender team according to Euclidean distance.
Performance improves in ADE, from 6.31 to 6.16 meters, and slightly degrades in terms of FDE, from 11.1 to 11.28. Overall
All distances are computed at the last observed frame. Furthermore, all distances are plain Euclidean distances, which a simple Neural Network may replicate or improve with E2E learning.

\textit{Distance from the ball and marking at future frames.}
The last row of Table~\ref{tab:oracle_orderings} considers the furthest future frame position for all distance computations.
It should be noted that the model makes no assumptions about future locations. Future information is simply utilized to place players in order.
This motivates us to replace the hand-defined ordering with an E2E-trained module, which we will do in the following section.

\input{src/tables/ordering.tex}

\subsection{End-to-end Model with latent roles}

In this section, we leverage the full RolFor model, E2E trained. Here the first module, OrderNN, sorts players into their roles in the action, then the RoleGCN module reasons on their role-based interaction. Sorting into roles has benefited forecasting in Sec.~\ref{sec:eucl-ord}. Here we assume that roles are latent variables, which the OrderNN estimates, E2E, based on the best forecasting performance.
Table~\ref{tab:e2e} compares the hand-defined baseline (ball and marking distance on the last observed frame, scoring 6.16/11.28) against E2E model variants. \textit{E2E} is learning to order, encode the role-role interaction, and forecast based on the encoder. This model is performing poorly at 12.12./15.02 ADE/FDE. Is this because the OrderNN is incapable of ordering, or is it because the OrderNN is not fully differentiable? 
Moreover, the \textit{EuclDistEst} variant attempts to answer part of this question. Here we used a pre-trained Neural Network module to approximate the Euclidean distance based on the player's performance. We then use the pre-trained module to sort players according to the ball. If the Euclidean distance estimator model were perfect, performance would be 6.31/11.1 (ADE/FDE), cf.~Table \ref{tab:oracle_orderings}. \textit{EuclDistEst} yields, however, 7.50/12.58. We attribute this mismatch to the residual errors in the Euclidean distance estimation, which, as it seems, matters. 
More surprisingly, \textit{E2E-finetune} starts from the \textit{EuclDistEst} variant, and it fine-tunes it, E2E. The error increases to 12.08/14.97, so the model neglects the initialization and reverts to the \textit{E2E} performance. We attribute the discrepancy between \textit{EuclDistEst} and \textit{E2E} to the challenges in the SoftRank differentiability, as we further analyze in the next section.

\input{src/tables/final_tabe.tex}

\subsubsection{Analysis of the Order Neural Network}

Here we focus on confirming our claims on the issues of the differentiability of Softrank. We set to order the players according to their ascending distance from the ball, at a specific frame, given their 2D coordinates. It allows us to test the first RolFor module, OrderNN, in isolation, cf.\ref{tab:ordering_results}.
In Table~\ref{tab:ordering_results}, we compare \textit{OrderNN E2E} against \textit{OrderNN EuclDistEst}. The first E2E trains the order of players and re-shuffles them. The second supervises the network by tasking it to learn the Euclidean distance between the players and the ball and then sort the distances according to SoftRank. 
We measure the ordering accuracy $p_{\text{ord}}$ as the percentage of players the models place in the correct order. In other words, we reproduce the top-k classification experiment as \cite{cuturi19}. The authors propose a loss for top-k classification between a ground truth class $ord \in [n]$ and a vector of soft ranks $\hat{ord} \in \mathbb{R}^n$, which is higher if the predicted soft ranks correctly place y in the top-k elements.

\input{src/tables/ordering_accuracy.tex}

Observe from Table~\ref{tab:ordering_results} that learning Euclidean distances from 2D positions is an easier task for a deep neural network since SoftRank yields 71\% at the \textit{top-10} ordering accuracy $p_{\text{ord}}$.
It is also interesting to notice that when changing the \textit{top-k} ordering accuracy into ${5, 3, 1}$, we get similar results to \cite{softrank}.
By contrast, learning the ordering E2E from the 2D coordinates yields surprisingly low performance. The table shows that \textit{OrderNN E2E} achieves a \textit{top-10} ordering accuracy of only 1\%.

\subsection{Robustness of RolFor to ordering errors}

How much does misordering impact forecasting? We measure ADE and FDE forecasting errors when randomly altering the order provided by our best performing oracle RolFor (5.55/9.99 ADE/FDE, Table \ref{tab:oracle_orderings}).
In more detail, we consider the swap of two players \textit{Light Swap}, which can occur if the distance between them is relatively small. A more significant error can also occur, e.g., one role is not identified correctly and a player is inserted at the wrong position, making the whole order slip. We name this \textit{Light Insert}. In Table~\ref{tab:swap_error}, we consider the two potential sources of errors by randomly simulating one or both.
The results are coherent with what we said previously Table \ref{tab:ordering_results}, where the RolFor \textit{EuclDistEst} has a \textit{top-10} ordering accuracy of $71\%$ yielding 7.50/12.58. At the same time, a Light Swap/Insert gives 6.55/12.10 in ADE/FDE and $80\%$ \textit{top-10} ordering accuracy. 
This last Table \ref{tab:swap_error} highlights the importance of roles and their impact on the final trajectory accuracy.

%% file: src/tables/comparison_existing.tex
\begin{table}[h!]
\centering
\caption{Comparison of our model with SoTA models}

\begin{tabular}{l|cc}

{Model} & ADE & FDE \\ \hline
EvolveGraph \cite{evolvegraph} & 5.73                              & \textbf{8.65}         \\ 
Social-STGCNN \cite{socialstgcn} & 6.42                              & 10.04        \\ 
STGAT \cite{stgat}          & 7.06                              & 12.54        \\ 
SGAN \cite{socialgan}  & 5.88                              & 10.36        \\ 
\textbf{RolFor} + Oracular Permutation  & \textbf{5.55}                              & 9.99        \\ 

\end{tabular}
\label{tab:generic_results}
\end{table}

%% file: src/tables/ordering.tex
\begin{table}[h!]
\centering
\caption{Results for different types of ordering}

\begin{tabular}{l|l|l|l|cc}
{Ball Dist.}&{Obs.}&{Future}&{Mark} & ADE  & FDE   \\ \hline
- & - & - & -  & 6.34 & 11.5 \\
\checkmark & - & - & - & 6.31 & 11.1 \\
\checkmark & \checkmark & -  & \checkmark & 6.16 & 11.28 \\
\checkmark & - & \checkmark & \checkmark & 5.55 & 9.99  \\ 
\end{tabular}

\label{tab:oracle_orderings}
\end{table}

%% file: src/tables/final_tabe.tex
\begin{table*}[h]
    \begin{subtable}[b]{0.45\textwidth}
        \centering
        \begin{tabular}[b]{l |ll}
        Configuration & ADE & FDE \\ \hline
        \textit{E2E} & 12.12 & 15.02 \\ 
        \textit{E2E-finetune} & 12.08 & 14.97 \\
        \textit{EuclDistEst} & 7.50 & 12.58 \\ 
        \textit{Best non-or. dist.} & 6.16 & 11.28 \\ 
        \end{tabular}
       \caption{Different training configurations for RolFor}
       \label{tab:e2e}
    \end{subtable}
    \hfill
    \begin{subtable}[b]{0.45\textwidth}
        \centering
        \begin{tabular}[b]{l|cc}
        {Model} & ADE & FDE   \\ \hline
        Oracular Ordering & 5.55                     & 9.99  \\
        Light Swap                            & 6.55                     & 12.10 \\
        Light Insert                          & 6.55                     & 12.10   \\
        Light Swap + Light Insert             & 6.59                     & 12.08 \\
        Heavy Swap + Heavy Insert             & 6.71                     & 12.25 \\ 
        \end{tabular}
        \caption{Analysis of simulated errors in ordering}
        \label{tab:swap_error}
     \end{subtable}
     \caption{Analysis on ADE and FDE for different approaches}
\end{table*}

%% file: src/tables/ordering_accuracy.tex
\begin{table}[h!]
    \centering
    \caption{ OrderNN \textit{E2E} against OrderNN \textit{EuclDistEst} top-k accuracy}
    
\begin{tabular}{ll|c}
        Model & top-k & Accuracy \\ \hline
        OrderNN \textit{E2E} & 10 & 1,00\% \\ 
        OrderNN \textit{EuclDistEst} & 10 & 71,00\% \\ 
        OrderNN \textit{EuclDistEst} & 5 & 77,00\% \\ 
        OrderNN \textit{EuclDistEst} & 3 & 82,00\% \\ 
        OrderNN \textit{EuclDistEst} & 1 & 92,00\% \\ 
    \end{tabular}
\label{tab:ordering_results}

\end{table}




%% file: src/sections/5.conclusion.tex
\section{Conclusions}

Our goal was to show that roles and social relations in sports are quantifiable and can be effectively used to improve the current SoA models in game forecasting.
We demonstrate that roles exist by testing different permutations over players. Then, we encode the player's coordinates into a latent space and use the encoding to find an optimal latent role ordering.
The model employed to perform trajectory forecasting is called RolFor (Role Forecasting) and considers the input nodes of a graph indicating roles in a game. This single-graph framework favors the relation between roles and time, allowing better learning of the fully-trainable adjacency matrices for role-role and time-time interactions. The adoption of CNNs and the graph structure of the input allows the requirement of parameters to be only a fraction of the ones used in Transformers, GANs, and VAEs. 
Our observations emphasize the significant opportunity for future work to develop fully differentiable ordering modules to enable learning latent role-based interactions in graph-based models, also applicable to social networks and multi-agent systems.

\paragraph{Acknowledgements}
This work was partially supported by the MUR PNRR project FAIR (PE00000013)